\documentclass[10pt,twocolumn,letterpaper]{article}

\usepackage{cvpr}              %

\usepackage[dvipsnames]{xcolor}

\usepackage{listings}
\usepackage{mdframed}
\definecolor{light-gray}{gray}{0.95}

\usepackage{microtype}
\usepackage[T1]{fontenc}
\usepackage[scaled]{beramono}
\newcommand\Small{\fontsize{8}{8.2}\selectfont}
\newcommand*\LSTfont{\Small\ttfamily\SetTracking{encoding=*}{-60}\lsstyle}

\graphicspath{{figures/}}
\graphicspath{{figures_supp/}}

\newcommand{\B}[1]{{\textbf{#1}}}
\newcommand{\SC}[1]{{\textsc{#1}}}

\newcommand{\refeqn}[1]{Eqn.~\ref{#1}}
\newcommand{\reffig}[1]{Fig.~\ref{#1}}
\newcommand{\reftbl}[1]{Table~\ref{#1}}
\newcommand{\refsec}[1]{Sec.~\ref{#1}}

\definecolor{cvprblue}{rgb}{0.21,0.49,0.74}
\usepackage[pagebackref,breaklinks,colorlinks,citecolor=cvprblue]{hyperref}

\def\paperID{10516} %
\def\confName{CVPR}
\def\confYear{2024}

\title{Step Differences in Instructional Video}

\author{Tushar Nagarajan, Lorenzo Torresani\\
FAIR, Meta\\
}

\begin{document}
\maketitle
\begin{abstract}
Comparing a user video to a reference how-to video is a key requirement for AR/VR technology delivering personalized assistance tailored to the user's progress. However, current approaches for language-based assistance can only answer questions about a single video. We propose an approach that first automatically generates large amounts of visual instruction tuning data involving pairs of videos from HowTo100M by leveraging existing step annotations and accompanying narrations, and then trains a video-conditioned language model to jointly reason across multiple raw videos. Our model achieves state-of-the-art performance at identifying differences between video pairs and ranking videos based on the severity of these differences, and shows promising ability to perform general reasoning over multiple videos. 
Project page: \url{https://github.com/facebookresearch/stepdiff}
\end{abstract}
\section{Introduction} \label{sec:intro}

Instructional \emph{how-to} videos are an important medium for learning new skills that offer in-depth visual demonstrations of complex procedural activities. In turn, they serve as a valuable resource for building AR/VR assistants that can guide a user through a procedural activity, by aligning user activity to a reference how-to video. Instructional videos have thus been the subject of several recent datasets and benchmarks that are driving new research~\cite{tang2019coin,zhukov2019cross,bansal2022my,htstep,goalstep,zhou2018towards,regneri2013grounding,sener2022assembly101}.  

A key requirement for such systems is the ability to compare and contrast the user's execution of a step in the activity with the reference video step, to highlight similarities and differences between them. For example, to let the user know that they used too much detergent (while doing laundry) or that the gravy is too thick (while cooking). This ability has direct value for personalized assistance applications such as progress tracking, mistake detection and surfacing user-activity driven tips. 

\begin{figure}[t]
\centering
\includegraphics[width=\linewidth]{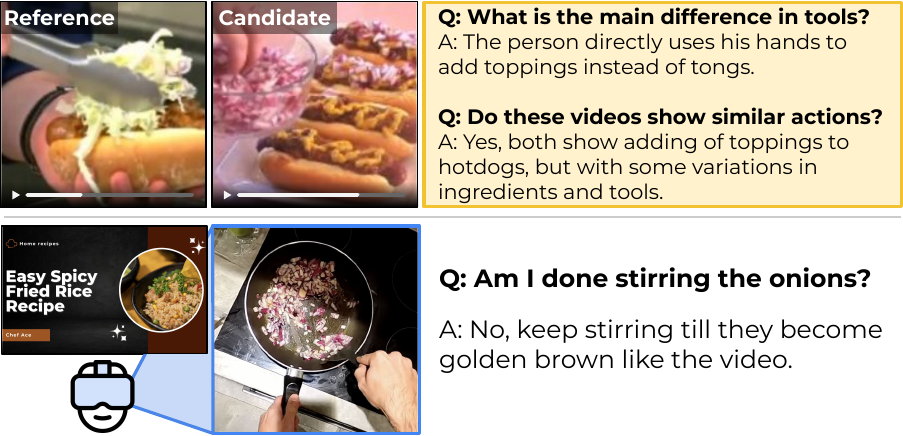}
\caption{\textbf{Main idea.} \textbf{Top:} We train models to compare two videos showing the same high-level keystep and to describe their differences (e.g., in tools, ingredients, technique). \textbf{Bottom:} Once trained, such models can then help answer questions about a user's activity compared to a reference (e.g., an internet how-to video) like ``did I do this step right?'' or ``am I done yet?''. 
}
\label{fig:concept}
\end{figure}

More generally, reasoning about a video with respect to a \emph{reference video} is a fundamental problem for video understanding that has value for fine-grained video retrieval~\cite{vo2019composing,wu2021fashion,goenka2022fashionvlp,ventura2023covr} (e.g., to browse internet videos for ``this movie scene, but in a forest''), step detection~\cite{sener2022assembly101,zhukov2019cross,sigurdsson2018actor,tang2019coin,bansal2022my} (e.g., to recognize subtle variations in keysteps) and multi-video question answering and reasoning~\cite{bansal2020visual,penamakuri2023answer} (e.g., to answer comparative questions like ``which video uses the least amount of oil?'').

Despite its importance, there has been limited work on comparing videos. Prior work has explored \emph{change captioning} in images~\cite{qiu2021describing,kim2021agnostic,park2019robust,zou2022spot,jhamtani2018learning,hosseinzadeh2021image}, however these works typically consider \emph{pixel-level} differences (e.g., missing or moved objects; changed background objects) in static scenes (e.g., the same parking lot; the same tabletop), or in synthetically generated datasets~\cite{park2019robust}. 
They do not consider important semantic differences in activities (e.g., differences in tool use, subtle variations in actions and techniques or visual differences due to state changes), which together with the low-level visual differences, form a complete picture of human-object interactions.

To address these limitations, we propose a video-conditioned language model (VCLM) approach to directly compare two videos of same step in a procedural activity. Specifically, we propose the \emph{difference question answering} task: given a reference and a candidate video, a model must answer a question that involves reasoning across both videos (e.g., what are the differences in tools? techniques?; do the two videos show the same activity?).
Such a model that effectively relates user activity to a reference video, can then provide detailed context to answer more general questions such as ``what did I do wrong compared to the reference'' or ``am I done yet?''. See \reffig{fig:concept}.

An important practical question is how to source the supervision to train such a model, given that existing video datasets only contain individual videos with captions. Moreover, meaningful differences are not guaranteed to exist between arbitrary pairings of videos. We therefore automatically generate training data from existing large-scale instructional video datasets annotated with keysteps and speech narrations describing what instructors are doing~\cite{miech2019howto100m,htstep}. 
We pair clips of the \emph{same keystep} (e.g., two videos of a person ``stir frying the rice until it is dark yellow'') but from distinct videos to allow for variations between them. For example the first video may use a cast iron pan versus a steel wok or the person may be tossing the food in the pan vs stirring with a spatula. We then leverage recent large language models~\cite{touvron2023llama} to generate questions and answers about the similarities and differences between the two videos given their visual descriptions, speech narrations, and visible objects as context. 
Inspired by work in visual instruction tuning of language models~\cite{liu2023visual,zhang2023instruction}, we finally fine-tune a \emph{video-pair} conditioned language model with the collected dataset. The resulting model has the ability to cross-reference videos to compare them, and more generally answer questions that require joint reasoning about both videos simultaneously. 

To evaluate our model, we collect a manually annotated dataset of 6292 video pairs with $\sim$36k difference captions spanning 5 categories, as well as scores for how severe the differences are. We set up the first benchmark for video comparisons and evaluate models on their ability to describe the differences in specific categories (e.g., ``What are the differences in tools? techniques?'') and to rank videos based on their differences (e.g., ``Which video shows the least different technique?''). 
Our models trained with weak-supervision from automatically generated data achieve state-of-the-art results on our benchmark, highlighting its value for personalized assistance applications. 
Our benchmark will be hosted publicly, to allow the community to make progress towards this under-explored task.

\section{Related work} \label{sec:related}

\paragraph{Instructional video understanding}
Recent large-scale instructional video datasets~\cite{tang2019coin,zhukov2019cross,bansal2022my,htstep,goalstep,zhou2018towards,regneri2013grounding,sener2022assembly101} have facilitated research in step captioning~\cite{zhou2018towards,zala2023hierarchical}, step detection~\cite{sener2022assembly101,zhukov2019cross,sigurdsson2018actor,tang2019coin,bansal2022my}, temporal grounding~\cite{anne2017localizing,bao2021dense,kuehne2016end,han2022temporal,mavroudi2023learning}, vision-language representation learning~\cite{pramanick2023egovlpv2,zhao2023learning,ashutosh2023hiervl,lin2022learning} and video question answering~\cite{xiao2021next,yu2019activitynet,wu2021star,yang2022learning} to name a few. In all these approaches, the goal is to process a single video and then caption, answer questions or temporally localize an action or text within it. While we are also interested in the space of procedural videos in the context of personalized language-based assistance, in contrast, we develop methods to compare and contrast \emph{multiple videos} --- namely a reference video and a candidate video --- in order to identify differences and answer comparative questions about them.

\paragraph{Visual differences in images}
Prior work has studied visual differences in images in the context of attributes~\cite{chen2018compare,parikh2011relative,yu2014fine,forbes2019neural} (e.g., which shoe is more formal) to facilitate fine-grained recognition. More relevant to our work, \emph{change captioning}~\cite{qiu2021describing,kim2021agnostic,park2019robust,zou2022spot,jhamtani2018learning,hosseinzadeh2021image} involves describing the differences between two images as a text caption. Other work defines differences as 2D bounding boxes~\cite{sachdeva2023changea,sachdeva2023changeb} or semantic maps~\cite{park2021changesim} for regions that differ. More recently, VCLM models have been trained with ``spot-the-difference'' data from the above with a similar goal of identifying image differences~\cite{li2023mimic}. In all these cases, the two images typically involve the same scene from multiple viewpoints or over time (e.g., surveillance footage) or are constructed from synthetic images (e.g., 3D geometric shapes re-arranged on a table). The resulting differences therefore focus on simple visual cues like missing or moved objects. More recent approaches use visual differences to retrieve videos~\cite{detours}, however they assume the difference is known (to retrieve a relevant video) rather than identifying and describing it.
In contrast, we compare across distinct video clips that show the same high-level keystep. As a result, the difference captions characterize complex variations that arise naturally from the availability of tools and ingredients, differing skill / technique or personal preference.

\paragraph{Visual instruction tuning of language models}
Given the recent success of large language models (LLMs), several efforts have tried to adapt them for use with various modalities including images, videos, audio etc., typically by aligning captions to modalities or instruction tuning~\cite{liu2023visual,zhang2023video,liu2023improved,li2023otter,gong2023multimodal,moon2023anymal,gpt4v,zhang2023instruction}.
All these approaches typically use text captions or generate instruction tuning data based on a \emph{single image or video}. In contrast, we generate instruction data for pairs of videos (a reference, and a target video) to allow vision conditioned language models to jointly reason about them both. Some approaches do train on multiple images interleaved with text~\cite{tsimpoukelli2021multimodal,alayrac2022flamingo,laurenccon2023obelisc}, however they do not support instructions at inference, and instead rely on in-context few-shot prompting to respond. In contrast, our approach can respond to arbitrary questions about a video with respect to a reference clip.

\begin{figure*}[t]
\centering
\includegraphics[width=\linewidth]{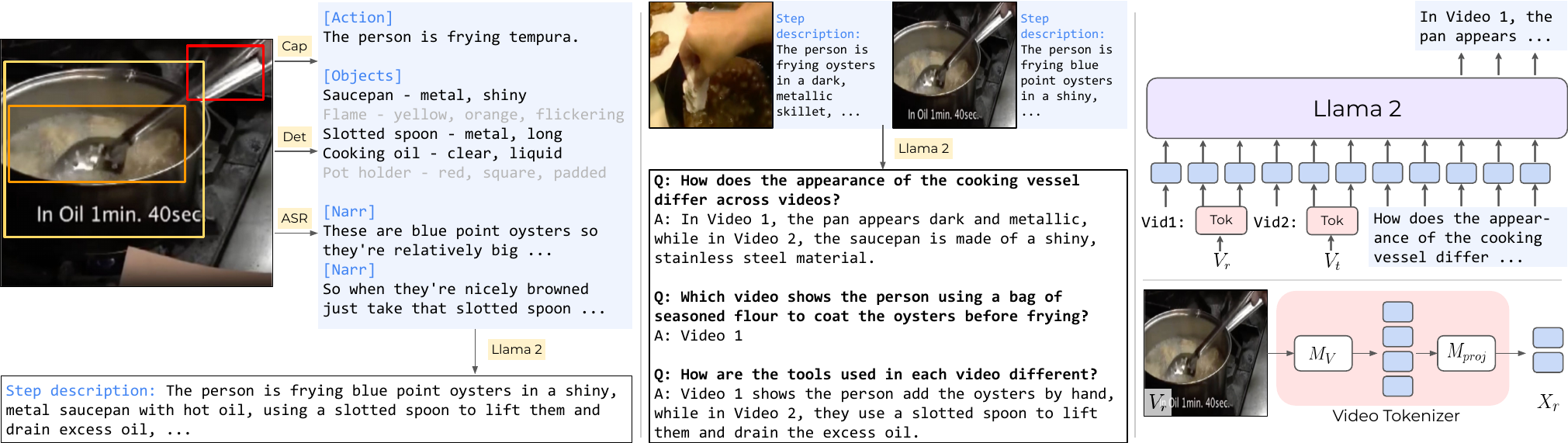}
\caption{\textbf{Step differences framework.} We first generate a comprehensive step description including information from action captions, object detections and ASR narrations (left panel). We then select pairs of clips with similar step descriptions, and automatically generate questions and answers that compare the two (center panel, \refsec{sec:dataset_gen}). Finally, we instruction-tune an LLM to generate answers conditioned on the generated questions and encoded representations of both videos (right panel, \refsec{sec:instruct_tuning}). Once trained, the model directly operates on video clips to compare them, without the need for captions, ASR or object detections.%
}
\label{fig:data_gen_pipeline}
\end{figure*}

\section{Approach} \label{sec:approach}
Our goal is to train models to answer questions about a video in the context of a reference video, by jointly reasoning about the two. The problem is two-fold: where do we source data of pairs of videos with relevant questions to train such models and what model architectures support training with multiple videos?
For the former, we turn to automatically generating this data using large-language models (LLMs) parsing narrated video from existing datasets. For the latter, we use vision-conditioned language models (VCLMs) --- a powerful class of models for single-video question answering --- adapted to our multi-video setting. In the following, we first formally define our task (\refsec{sec:task_def}). Next, we describe our automatic training data generation pipeline (\refsec{sec:dataset_gen}). Finally, we discuss training and downstream inference (\refsec{sec:instruct_tuning}).

\subsection{Task definition} \label{sec:task_def}
We require models that collectively answer questions about two videos. Formally, given a reference video $V_r$, a candidate video $V_c$ and a question $q$, models must produce a corresponding answer $a$. This formulation is an extension of standard video question answering or captioning~\cite{zhong2022video} with a response that is additionally conditioned on a reference video. The questions can take various forms, for example ``How is the dough being prepared differently in Video 2''; ``What is the similarity in mixing techniques between the two videos?''. Critically, these questions all share the assumption that a single video alone (either the reference or the candidate) is insufficient to answer the question --- reasoning over both videos is required. 

In our experiments, we train models with a diverse set of automatically-generated question-answer pairs. At test time, we focus on \emph{step differences}, where the $q$ is of the form ``what is the main difference between these videos in the category $g$'' and $g$ is the difference category (e.g., ingredients, techniques, etc.). This structure captures a representative range of fine-grained differences, and allows for consistent evaluation of models as we will show.

\subsection{Step differences dataset generation} \label{sec:dataset_gen}
To train our models, we require a dataset of paired videos along with questions and answers (QA) relating the two in the form $(V_r, V_c, q, a)$. However, current video datasets typically contain individual video clips annotated for actions, narrations, or single-video QA which is incompatible with our task definition. We therefore construct this from existing video datasets using large-language models, inspired by prior work on instruction tuning~\cite{taori2023alpaca,liu2023visual,li2023mimic,gong2023multimodal,peng2023instruction}.

Constructing this dataset from existing video datasets is non-trivial. On the one hand, selecting random pairs of videos showing very different content (e.g., sports vs. cooking) or near-identical videos (e.g., from repetitions of the same activity by the same participant) will lead to trivial differences. On the other hand, naively selecting video pairs of the same class in action recognition datasets (e.g., ``Bookbinding'' or ``Mowing the lawn'') will not highlight fine-grained differences of interest, and will instead focus on global differences (e.g., changes in actors or scenes). Moreover, these datasets do not come with text descriptions to construct differences from.

We therefore propose to use videos from the large-scale procedural video dataset HowTo100M~\cite{miech2019howto100m}, specifically cooking-themed videos labeled for keysteps from HT-Step~\cite{htstep}. Instructional videos are an ideal data source as they are narrated and show the same high-level keystep, but with variations that arise naturally from availability of tools and ingredients, differing skill / technique or personal preference. 

Specifically, for two videos showing the same keystep (e.g., Slowly pour the sauce over the dumplings), we assume one is the reference $V_r$ and the other is the candidate video $V_c$, with corresponding speech narrations. First, we generate descriptions of the actions and objects (including their attributes) using off-the-shelf captioning models~\cite{moon2023anymal}. These models often hallucinate details in their generations, so we additionally filter object descriptions based on the scores of a pre-trained detection model~\cite{minderer2023scaling} and filter action descriptions using visual grounding models~\cite{wang2022internvideo}. Details about the filtering stage are in Supp. %
Finally, we aggregate the information from these three sources (ASR narration, filtered objects and actions) to synthesize a detailed step description for each video (\reffig{fig:data_gen_pipeline}, left panel). We then prompt a language model (in our case, Llama 2~\cite{touvron2023llama}) to generate both questions and answers comparing the two videos based on their step descriptions. In short, the prompt takes the form: ``Video 1: \{description1\}. Video 2: \{description2\}. Summarize the differences and generate 3 question-answer pairs comparing the two videos.'' (\reffig{fig:data_gen_pipeline}, center panel).
An overview of the data generation pipeline with examples at each stage can be seen in \reffig{fig:data_gen_pipeline}. See Supp. for more examples and full step description prompt details.

The resulting dataset contains QA instances over video pairs across 87740 unique video clips. Note that the LLM-generated data is noisy --- they may hallucinate details that are not present in the video, misunderstand the ASR narrations, produce irrelevant questions or incorrect answers to questions. Despite this, they offer valuable \emph{weak supervision} to train our VCLM models, as our experiments will show. 

\subsection{Paired video instruction tuning} \label{sec:instruct_tuning}
We require a model that can generate natural language responses to video comparison questions in our dataset. To do this, we adapt a vision-conditioned language model (VCLM) to our multi-video setting via visual instruction tuning. In short, visual instruction tuning aligns the outputs of an image (or video) backbone to a powerful LLM to condition its responses on the visual content. This strategy has been successful in prior work for single image/video captioning and question answering~\cite{liu2023visual,zhang2023video,liu2023improved,li2023otter,gong2023multimodal,moon2023anymal,zhang2023instruction}. We extend this to support comparisons across multiple videos. In our experiments, we use a Llama2~\cite{touvron2023llama} LLM aligned with an Internvideo~\cite{wang2022internvideo} backbone following prior work~\cite{moon2023anymal}. 
Note that it is possible to directly provide multiple videos to existing models by adding extra visual tokens to the input prompt, however their performance is degraded as they not trained to support this. We compare against such models.

Specifically, for an instruction-tuning instance $(V_r, V_c, q, a)$, we generate an instruction prompt in the Llama2 format as follows.
\begin{mdframed}[backgroundcolor=light-gray, roundcorner=10pt,leftmargin=0, rightmargin=0, innerleftmargin=4, innertopmargin=0, innerbottommargin=0, outerlinewidth=0, linecolor=light-gray]
\begin{lstlisting}[basicstyle=\LSTfont, breaklines=true, breakindent=0pt]
<s> [INST] <<SYS>> You are a helpful AI assistant that answers questions about a pair of videos. Answer in a single sentence. Here is the first video: {V_r}. Here is the second video: {V_c}. 
<</SYS>> {q} [/INST] {a}
\end{lstlisting}
\end{mdframed} 

We encode the text tokens in this prompt using the LLM's pre-trained text encoder. We encode each video into a sequence of spatiotemporal tokens using a pre-trained video backbone $M_{V}$, and then align them to the LLM's input space using a learnable projection module $M_{proj}$. The resulting encoded instruction prompt is a sequence of tokens comprising a mix of text and visual tokens, which can then be processed by the LLM (\reffig{fig:data_gen_pipeline}, right panel). 

The model is trained using the original auto-regressive objective to maximize the probability of generating the answer tokens, conditioned on the question, reference and candidate video, and is trained using a standard cross-entropy loss.

\begin{align}
    p(X_a | X_r, X_c, X_q) &= \prod_{i=1}^{|X_a|} p_{\theta}(X_{a,i}|X_r, X_c, X_q, X_{a,<i}) \\ \label{eq:likelihood}
    X_{r} &= M_{proj}(M_{V}(V_{r}))\\  
    X_{c} &= M_{proj}(M_{V}(V_{c})),
\end{align}
where $X_{a,i}$ 
is the i-th answer token in the sequence, $X_{r}$ ($X_{c}$) are the visual tokens corresponding to the reference (candidate) videos, $X_q, X_a$ are tokens of the question and answer, $X_{a,<i}$ are answer tokens that occur before $X_{a,i}$ and $\theta$ are the learnable parameters in $\mathcal{M}_{proj}$.

Note that the video encoder and the LLM weights are frozen, and the loss is computed only for answer tokens. Only the projection layer is fine-tuned. Once trained, our model will be able to refer to each video, discuss their similarities and compare them. We evaluate our model by autoregressively generating text in response to various prompts coupled with reference and candidate videos.

\begin{figure}[t]
\centering
\includegraphics[width=\linewidth]{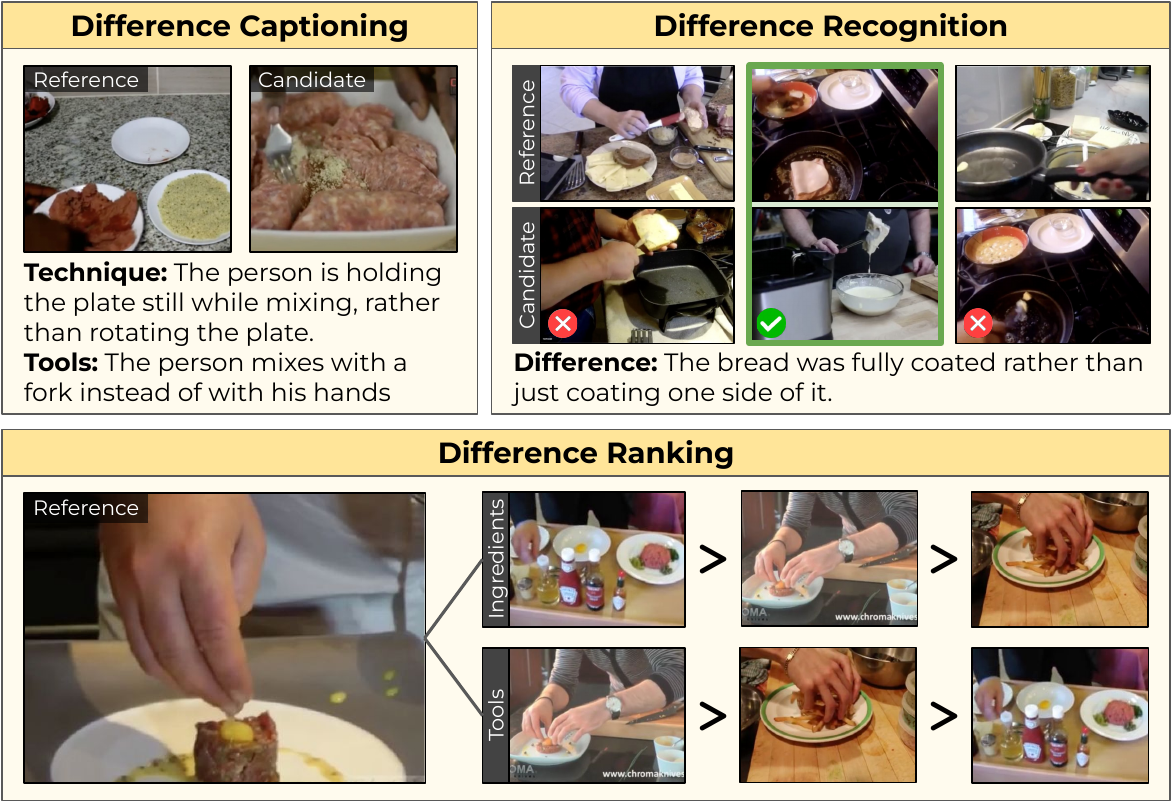}
\caption{\textbf{Evaluation tasks.} We evaluate on describing (DiffCap), recognizing (DiffMCQ) and ranking (DiffRank) differences.
}
\label{fig:tasks}
\end{figure}

\begin{figure*}[t]
\centering
\includegraphics[width=\linewidth]{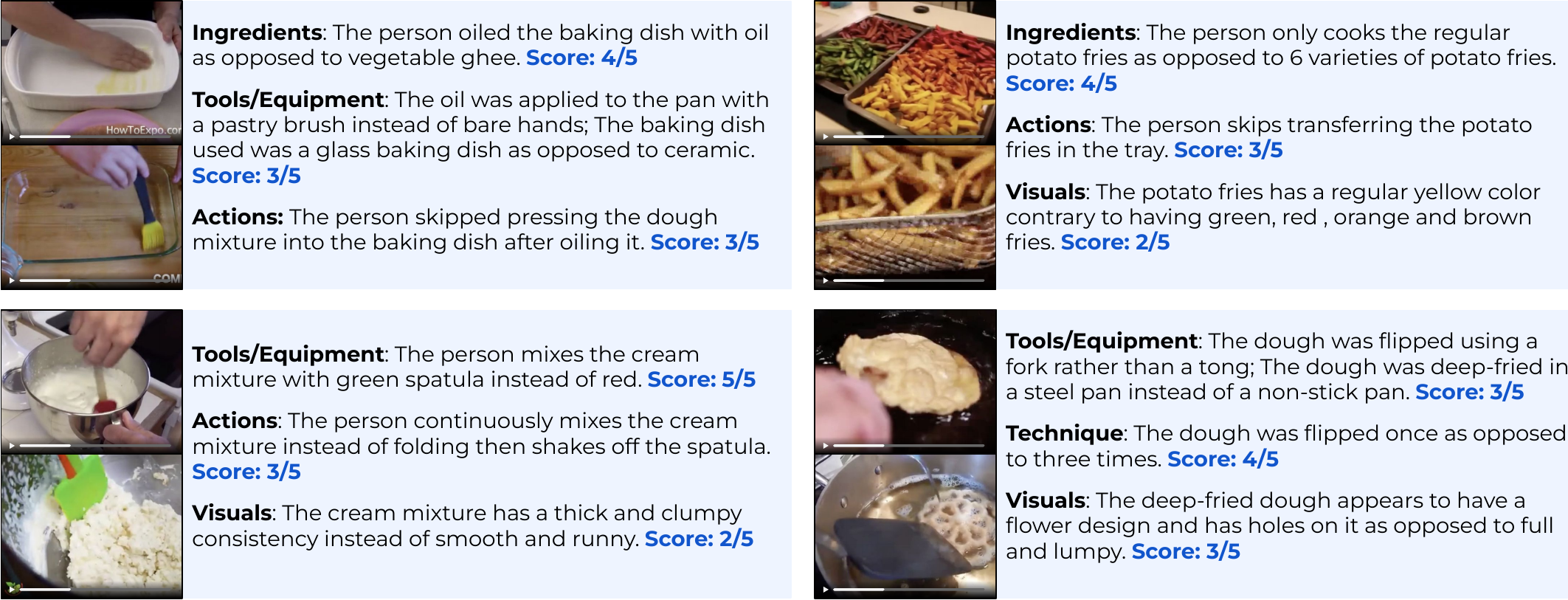}
\vspace{-0.2in}
\caption{\textbf{StepDiff dataset samples} We annotate text describing differences in various categories and scores for \emph{how different} the videos are in each category (1 = very different; 5 = nearly identical). More examples are in Supp. 
}
\vspace{-0.1in}
\label{fig:annot_examples}
\end{figure*}

\subsection{Describing, recognizing and ranking step differences in procedural videos} \label{sec:downstream}

Finally, we use our trained models to identify and rank fine-grained differences between pairs of video. We cast these tasks into the paired-video QA framework as follows.

\vspace{-0.05in}
\paragraph{Difference captioning (DiffCap)} The goal is to generate a textual description of the differences between two videos in a specific category $g$ (e.g., ingredients, tools). The question $q$ takes the form ``what is the main difference between these videos in the category $g$''. The difference caption is generated auto-regressively using the trained model.

\vspace{-0.05in}
\paragraph{Difference recognition (DiffMCQ)} The goal is to select the correct video pair that matches the difference caption, from a list of candidate video pairs $\{(V_r^i, V_c^i)\}_{i=1..4}$. This is a discriminative version of the captioning task above inspired by recent work in vision-language feature learning~\cite{lin2022egocentric}. For this, we compute $p(a | V_r^i, V_c^i, q)$ --- the likelihood of generating the difference text given the pair of videos following \refeqn{eq:likelihood} ---  and then select the pair with the highest score. 

\vspace{-0.05in}
\paragraph{Difference ranking (DiffRank)} The goal is to rank video instances $\{V_c^i\}_{i=1..4}$ based on how different they are to a common reference video $V_r$, in terms of a particular category of interest $g$. For this, we set $q$ to be ``do these two videos show the same $g$? Answer YES or NO.'', and rank each candidate video based on the likelihood of generating ``YES'' as the response.  

Together, these tasks are a representative suite of problems for instructional video understanding that require comparing videos along various axes. DiffCap tests how accurately a model can describe differences in natural language, DiffMCQ tests how well it can discriminate differences between videos, and DiffRank tests how well the model can assess the severity of these differences to rank them. A model for these tasks can enable applications that guide user action (e.g., to follow a reference video tutorial) or help browse through large collections of videos (e.g., to find the perfect variation of a recipe). \reffig{fig:tasks} illustrates these tasks.

\section{Experiments} \label{sec:experiments}
We evaluate our VCLM model on the three step difference tasks from \refsec{sec:downstream}. %

\begin{table*}[]
\centering
\small
\begin{tabular}{|l|ccc|c|c|c|c|}
\multicolumn{1}{c}{} & \multicolumn{3}{c}{\SC{DiffCap}} & \multicolumn{1}{c}{} & \multicolumn{1}{c}{\SC{DiffMCQ}} & \multicolumn{1}{c}{} & \multicolumn{1}{c}{\SC{DiffRank}} \\
\cline{2-4}\cline{6-6}\cline{8-8}
\multicolumn{1}{c|}{}
& BLEU      & CIDER     & ROGUE-L   & & Acc \%      & & $\tau$      \\ \cline{1-4}\cline{6-6}\cline{8-8}
VLEmbed (CLIP)~\cite{radford2021learning}
& --        & --        & --        & & 0.396       & & 0.127       \\ 
VLEmbed (InternVideo)~\cite{wang2022internvideo}
& --        & --        & --        & & 0.451       & & 0.170       \\ \cline{1-4}\cline{6-6}\cline{8-8}
Socratic (BLIP-2)~\cite{li2023blip}
& 0.159     & 0.036     & 0.169     & & 0.335       & & 0.022       \\
Socratic (LLaVA)~\cite{liu2023visual}
& 0.151     & 0.031     & 0.166     & & 0.332       & & 0.004       \\
Socratic (Step desc.)
& 0.141     & 0.020     & 0.172     & & 0.392       & & 0.013       \\ \cline{1-4}\cline{6-6}\cline{8-8}
VCLM (LLaVA)~\cite{liu2023visual}
& 0.211     & 0.069     & 0.199     & & 0.381       & & 0.019       \\
VCLM (AnyMAL)~\cite{moon2023anymal}
& 0.209     & \B{0.115} & 0.196     & & 0.471       & & 0.032       \\
Interleaved (IDEFICS)~\cite{laurenccon2023obelisc}
& 0.217	    & 0.080	    & 0.210     & & 0.376       & & 0.081       \\
Interleaved (AnyMAL)
& 0.207     & 0.102     & 0.198     & & 0.497       & & 0.014       \\ \cline{1-4}\cline{6-6}\cline{8-8}
StepDiff
& \B{0.223} & 0.104     & \B{0.215} & & \B{0.541}   & & \B{0.181}   \\ \cline{1-4}\cline{6-6}\cline{8-8}
\end{tabular}
\vspace{-0.1in}
\caption{\textbf{Results.} Our approach outperforms three classes of baselines built on top of state-of-the-art vision-language embedding and VCLM models. VLEmbed baselines are excluded from DiffCap as they cannot generate text.}
\label{tbl:results}
\end{table*}

\paragraph{Dataset} We construct a test dataset from videos in HTStep~\cite{htstep}. HTStep contains videos from a large-scale procedural video dataset, HowTo100M~\cite{miech2019howto100m} (\emph{Cooking \& Entertainment}), with temporal segments (clips) annotated for keysteps (e.g., ``fry then onions until golden brown''). 
We manually annotate pairs of clips, where each pair corresponds to instances of the same labeled keystep, but from distinct videos. Annotators are asked to identify the main differences across 5 categories (ingredients, tools/equipment, techniques, visual differences) and write difference captions of a consistent style --- what happens in the target clip, compared to what happens in the reference (e.g., ``The person uses a deep fryer to fry the potatoes instead of shallow frying it in a pan''). They are then asked to score the difference caption in each category on a scale of 1-5 based on how severe the difference is, where 1 is a significant difference (e.g., swapping out a critical ingredient that would change the dish entirely) and 5 is nearly identical (e.g., minor cosmetic differences that does not affect the activity). A rubric is used to ensure consistency in scoring. 

Note that this data is only used for evaluation --- we exclude these pairs from the automatic training data generation pipeline described in \refsec{sec:dataset_gen} to ensure that the model has not seen these instances during training. In total, we collect 35988 difference captions across 6292 clip pairs, involving 8396 unique clips. See \reffig{fig:annot_examples} for examples. Full collection details and dataset statistics are in Supp.

\vspace{-0.05in}
\paragraph{Baselines} We compare several classes of models. 
\begin{itemize}[leftmargin=*]
\itemsep0em 
\item \B{VLEmbed} is a class of vision-language model that embeds images or video in the same space as text, and then compares their similarity in the shared space. Video \emph{pair} embeddings are calculated as the average of individual video embeddings\footnote{We evaluate other aggregation strategies in Supp.}. We use CLIP~\cite{radford2021learning} and InternVideo~\cite{wang2022internvideo}. %
\item \B{Socratic} is a class of VCLMs that first converts videos into text using a captioning model, and then prompts a text-only LLM with these captions. These models are powerful, but often require complex, manually engineered prompts. We use state-of-the-art visual captioners (BLIP-2~\cite{li2023blip}, LLaVA-1.5~\cite{liu2023visual}) as well our aggregate step descriptions from \refsec{sec:dataset_gen}. We use Llama2 to process the captions regardless of which model generated them, for fair comparisons.
\item \B{VCLM} is a class of visual instruction-tuned language model trained for video captioning and question answering (for a single video). We directly add extra tokens for the reference video into the prompt to be consistent with our paired-video QA task. We compare LLaVA-1.5~\cite{liu2023visual} and AnyMAL~\cite{moon2023anymal}.
\item \B{Interleaved} is a class of models that are trained with interleaved sequences of images/videos and text, and naturally support multiple videos as inputs, but are not explicitly trained to compare them. We compare the recently proposed IDEFICS~\cite{laurenccon2023obelisc} and a model we train on sequences of (video, ASR) pairs from HowTo100M (training details in Supp.).
\end{itemize}

These baselines represent a spectrum of leading strategies for vision-language reasoning, including methods that directly embed video and language in the same space (VLEmbed), ones that explicitly convert videos to text and perform exclusively text-based reasoning (Socratic) and ones that perform joint vision-text reasoning on videos (VCLM, Interleaved). We ensure that each class of baselines include methods that have been trained on in-domain HowTo100M videos, while excluding the evaluation videos, to ensure fair comparisons with our approach. These are InternVideo, Socratic (Step desc.), VCLM (AnyMAL), and Interleaved (AnyMAL). Additional pretraining and implementation details are in Supp.

\vspace{-0.05in}
\paragraph{Implementation details} We use the Llama2-chat-70B~\cite{touvron2023llama} as the base LLM for all our experiments. Following prior work~\cite{moon2023anymal}, $M_V$ is an Internvideo~\cite{wang2022internvideo} video encoder that inputs 8 uniformly sampled frames from each video clip and generates 2056 spatio-temporal tokens. $M_{Proj}$ is a 2-layer Perceiver~\cite{jaegle2021perceiver} module followed by a linear layer head to output 32 tokens in the LLM's input dimension. %
During training, all parameters are frozen except for $M_{Proj}$. StepDiff models are initialized from Interleaved model weights before finetuning (interleaved data is retained during finetuning). For baselines, we use the largest available versions of models --- InstructBLIP (Vicuna13B), LLaVA (Vicuna13B), AnyMAL (70B), IDEFICS (80B). Full implementation and training details are in Supp.

\begin{figure*}[t]
\centering
\includegraphics[width=\linewidth]{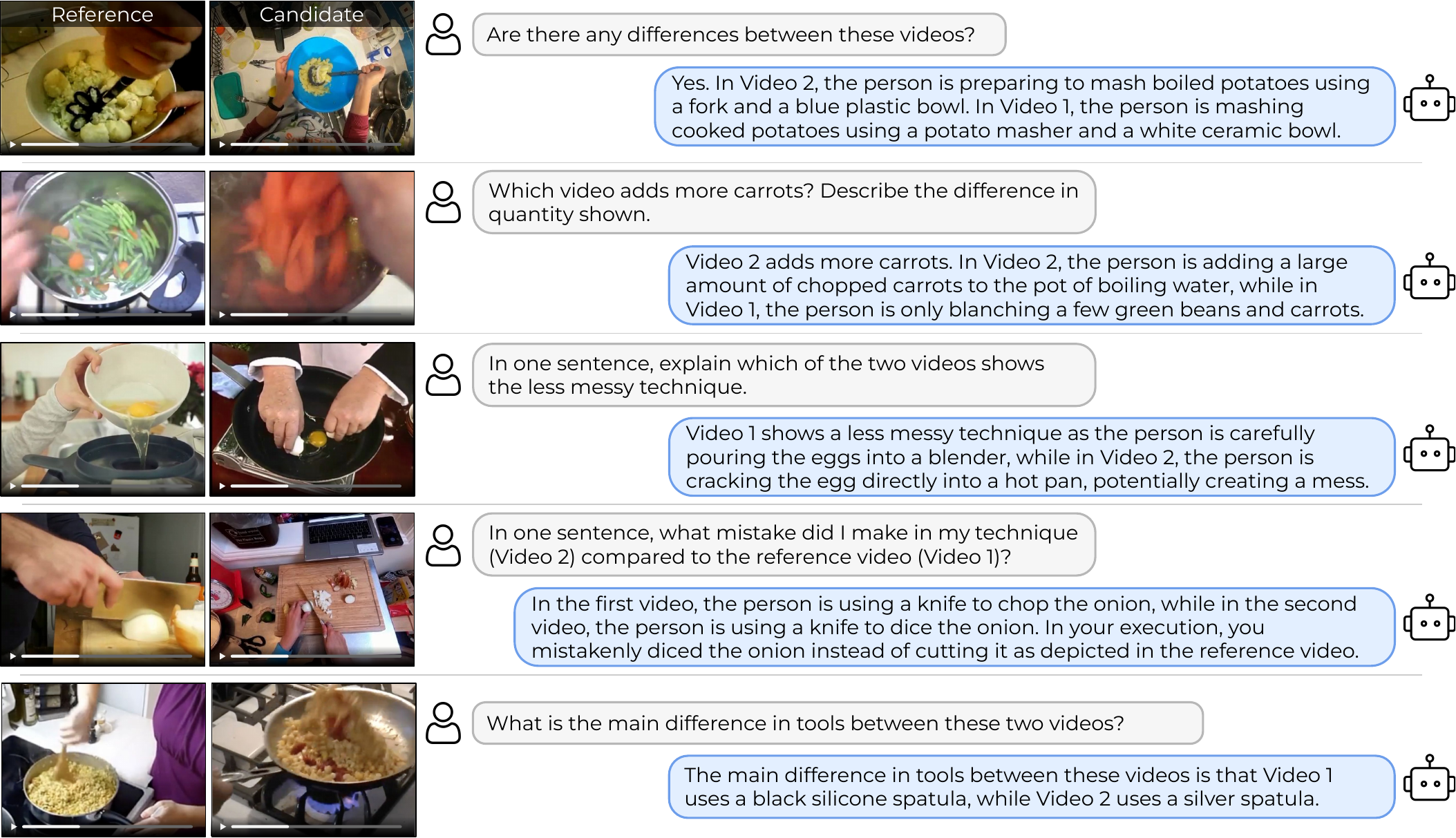}
\vspace{-0.1in}
\caption{\textbf{Extended QA on video pairs.} Our model which can describe differences (row 1) can be prompted (i.e., queried without any form of retraining) for comparative reasoning (e.g., ``why are they different?'', ``how different are they?'' row 2-3), or to bootstrap mistake detection (row 4). A failure case is shown in row 5 due to model hallucination.
}
\vspace{-0.1in}
\label{fig:extended_qa}
\end{figure*}

\subsection{Difference captioning} \label{sec:diffcap}
\begin{figure}[t]
\centering
\includegraphics[width=\linewidth]{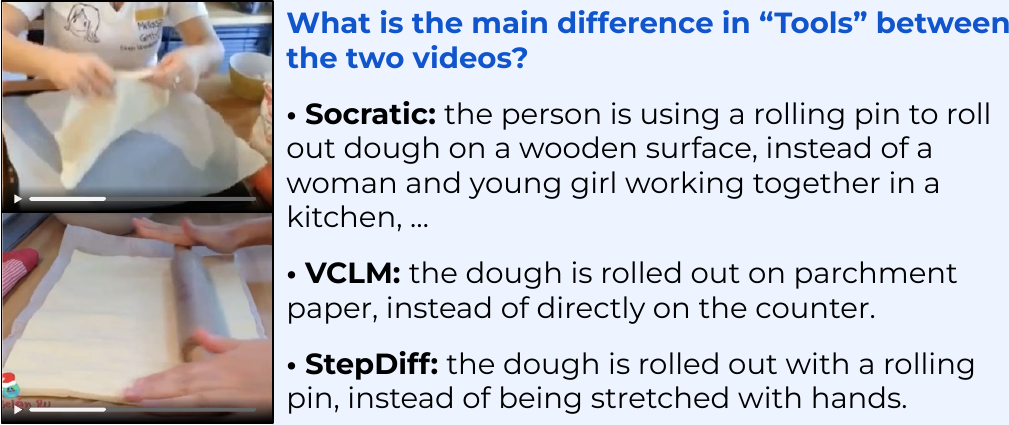}
\vspace{-0.1in}
\caption{\textbf{DiffCap baselines.} Our approach can describe differences without relying on input captions (like Socratic) and is less prone to  hallucinating details (like VCLM).
}
\label{fig:diffcap_results}
\vspace{-0.1in}
\end{figure}

We first evaluate how well our model can describe differences in video pairs (DiffCap). As mentioned in \refsec{sec:task_def}, $q$ is of the form ``what is the main difference between these videos in the category $g$'' where $g$ is the difference category. Since there may be multiple annotated differences in the same category, we group them together and treat them as a ground truth set, %
resulting in a dataset with 22292 instances.
We measure standard text generation metrics including CIDER~\cite{vedantam2015cider} and ROUGE-L~\cite{lin2004automatic}. Outputs are post-processed using simple string matching techniques to ensure difference captions are generated in the correct format (details in Supp). For the socratic baselines, we provide the generated caption instead of the video tokens in the prompt from \refsec{sec:instruct_tuning}. \reftbl{tbl:results} (left) shows our results. 
The socratic models perform poorly as they are limited by the information contained in the base captions. It is infeasible to generate captions that exhaustively describe every aspect of a video, without knowing what is of interest, and without the risk of model hallucinations. The VCLM models perform better, especially when trained to process multiple videos (i.e., interleaved models), however they still fall short of our approach that can explicitly compare and contrast videos.
The example in \reffig{fig:diffcap_results} highlights the sensitivity of socratic models to input captions (e.g., the reference caption did not mention the use of hands), and shows how VCLM models tend to hallucinate details. Our approach can correctly describe the difference. See Supp for more examples. %

\subsection{Difference recognition} \label{sec:diffmcq}
\begin{figure}[t]
\centering
\includegraphics[width=\linewidth]{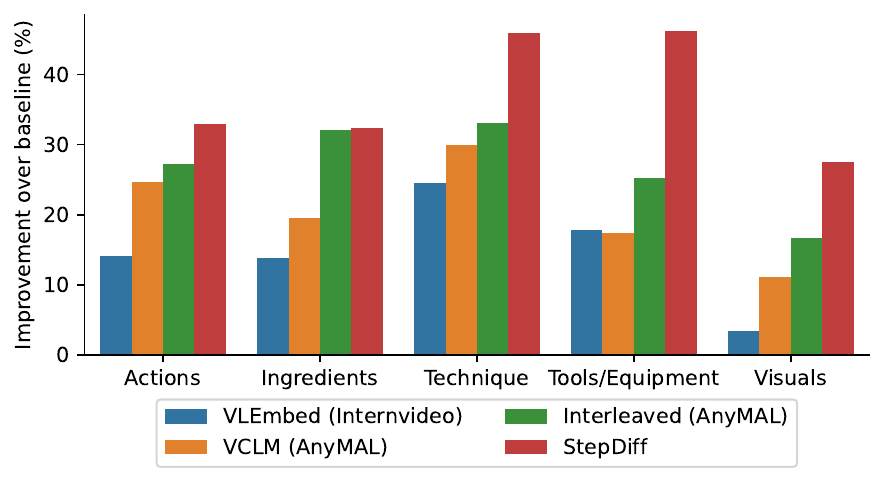}
\vspace{-0.1in}
\caption{\textbf{DiffMCQ performance by category.} Improvements are reported over the weakest baseline (Socratic).}
\label{fig:diffmcq_perclass}
\vspace{-0.1in}
\end{figure}

While the captioning metrics are informative, they are based on word overlap statistics, and do not always capture the semantics of the text well. To address this, we evaluate on DiffMCQ -- the discriminative version of the captioning task.  We adapt the same dataset from DiffCap, except we sample a single difference caption for each category if there are multiple differences present. Further, we sample three \emph{negative} video pairs from other instances in the dataset that involve similar objects and actions (details in Supp).
For the VLEmbed baselines, we score each video pair using the cosine similarity between their average visual embeddings and the text embedding of the difference caption. We compare variants of this baseline considering only the reference or target in Supp. For all LLM-based baselines, we compute the likelihood of generating the difference caption for each video pair, under each model as discussed in \refsec{sec:task_def}. We evaluate top-1 accuracy.
\reftbl{tbl:results} (center) shows our results. The joint feature embedding models capture some semantics, but are insufficient for identifying differences. Socratic models have a similar trend to captioning results, however models trained on step differences show large improvements, highlighting the value of careful curation for generating captions. Among VCLM models, ones that have seen in-domain HowTo100M videos during training have an edge over the others (i.e., LLaVA, IDEFICS), with interleaved models again being superior. Our model outperforms all these approaches with a 5\% accuracy improvement over the strongest baseline. \reffig{fig:diffmcq_perclass} shows performance increases by difference category, over the weakest baseline (Socratic). Our approach shows large relative improvements on most categories especially in technique and tool use (both 46\%), which require fine-grained action understanding.

\subsection{Difference ranking} \label{sec:diffrank}
Finally, we evaluate how well our model can rank videos based on the severity of differences compared to a common reference (DiffRank). Each reference video in the dataset is paired with four target videos, scored along each category axis. For example, some videos may be very similar in terms of technique, but very different in terms of ingredients. We only retain instances where there is a clear ranking (i.e., no more than one tie in scores) The resulting dataset contains 3932 instances involving 5746 unique clips.

As discussed in \refsec{sec:task_def}, we rank each target video candidate based on the likelihood of producing the response ``YES'' when asked whether it is similar to the reference. We use the Kendall's $\tau$ rank correlation metric to evaluate how well the generated ranking compares to the ground truth ranking annotators provide.
\reftbl{tbl:results} (right) shows our results. Unlike the previous tasks, the joint embedding models perform better than the LLM based baselines for two reasons. First, similarity in the embedding space directly translates to a score for ranking, rather than relying on computing ``YES'' token probability as a proxy for this. Second, there is a high correlation between the rankings across categories for the same set of instances ($\tau=0.63$). For example, videos ranked low in similarity for \emph{tools} when compared to a reference are often also ranked low for \emph{technique}. Despite these issues, our approach is able to outperform all baselines, showcasing its versatility as a retrieval and ranking model.

\subsection{Extending QA beyond atomic differences} \label{sec:extended_qa}
Next, we show how our model can be prompted to answer questions beyond just ``describe the differences''. LLMs have shown remarkable abilities for complex, multi-step reasoning in text -- our training framework unlocks the same kind of reasoning for multiple videos, based on their differences. In \reffig{fig:extended_qa}, we show some examples of this. Our model is able to naturally describe differences as it was trained for this task (row 1), but also has the ability to perform comparative reasoning (row 2-3) or explain mistakes (row 4). We show a failure case in row 5, where the model hallucinates content -- a characteristic feature of the LLM models it is built upon. Moreover, our model works with egocentric video (row 1, 4), despite being trained on largely third-person video content (HowTo100M), which is promising for AR/VR user assistance applications.

\begin{table}[]
\centering
\resizebox{\columnwidth}{!}{%
\begin{tabular}{|l|ccc|c|c|c|c|}
\multicolumn{1}{c}{} & \multicolumn{3}{c}{\SC{DiffCap}} & \multicolumn{1}{c}{} & \multicolumn{1}{c}{\SC{DiffMCQ}} & \multicolumn{1}{c}{} & \multicolumn{1}{c}{\SC{DiffRank}} \\
\cline{2-4}\cline{6-6}\cline{8-8}
\multicolumn{1}{c|}{}
& BLEU      & CIDER     & ROGUE-L   & & Acc \%      & & $\tau$      \\ \cline{1-4}\cline{6-6}\cline{8-8}
StepDiff
& \B{0.223} & 0.104     & \B{0.215} & & \B{0.541}   & & 0.181        \\
w/o interleaved data  
& 0.214     & 0.094     & 0.210     & & 0.499       & & \B{0.185}      \\ 
w/o QA filtering 
& 0.222     & 0.096     &  0.212    & & 0.516       & & 0.120       \\ 
w/ 13B LLM 
& 0.216     & \B{0.124} & 0.205     & & 0.527       & & 0.175       \\ \cline{1-4}\cline{6-6}\cline{8-8}
\end{tabular}
}
\vspace{-0.1in}
\caption{\textbf{Ablation experiments.} Impact of retaining interleaved training data, careful filtering of QA training data and LLM size.}
\label{tbl:ablations}
\vspace{-0.1in}
\end{table}

\subsection{Ablation experiments} \label{sec:ablations}
Finally, we ablate several design choices in our model in \reftbl{tbl:ablations}. As mentioned in \refsec{sec:experiments}, we finetune models on both interleaved ASR data as well as our generated pair QA data. Without the interleaved data, the model performance drops on two tasks, likely due to catastrophic forgetting (\emph{w/o interleaved data}). Next, we show the importance of filtering the generated QA data (\emph{w/o QA filtering)}, given the high likelihood of hallucinations produced by the LLM. Finally, we swap out the 70B LLM model for a smaller sized one (\emph{w/ 13B LLM}), causing the performance to drop, though not significantly.

\section{Conclusion} \label{sec:conclusion}
We proposed StepDiff, a video-conditioned language model (VCLM) that can compare and contrast videos to reveal fine-grained differences between them. We propose an approach that can automatically generate instruction-following paired-video QA training data from large-scale procedural video data, and a manually curated benchmark to evaluate models. Our experiments on describing and identifying differences, as well on ranking videos based on differences demonstrate the value of our approach for personalized assistance applications. Future work can leverage our work for personalized retrieval (e.g., retrieve content based on user-activity), or multi-video QA beyond instructional videos.

\vspace{0.1in}
\noindent \textbf{Acknowledgements} Thanks to Efi Mavroudi, Huiyu Wang, Triantafyllos Afouras and Yale Song for helpful discussions; Kumar Ashutosh and Suyog Jain for help with annotation tooling and collection; Austin Miller and Honey Manglani for managing the annotator workforce.

{
    \small
    \bibliographystyle{ieeenat_fullname}
    \bibliography{main}
}

\clearpage
\setcounter{page}{1}
\maketitlesupplementary

\setcounter{section}{0}
\setcounter{figure}{0}
\setcounter{table}{0}
\renewcommand{\thesection}{S\arabic{section}}
\renewcommand{\thetable}{S\arabic{table}}
\renewcommand{\thefigure}{S\arabic{figure}}

This section contains supplementary material to support the main paper. The contents include:

\begin{itemize}[leftmargin=*]
\itemsep0em 
    \item (\ref{sec:supp_data_gen}) Training data generation details, including full prompts, description of data filtering implementation and additional examples to supplement \refsec{sec:dataset_gen}.
    \item (\ref{sec:supp_annotation_details}) Annotation collection details and dataset analysis to supplement \refsec{sec:experiments} (dataset) and \reffig{fig:annot_examples}.
    \item (\ref{sec:supp_implementation_details}) Full implementation and training details for baselines and our approach to supplement \refsec{sec:experiments}.
    \item (\ref{sec:supp_tasks}) Additional task formulation details including post-processing implementation for DiffCap (\refsec{sec:diffcap}) and DiffMCQ negative sampling (\refsec{sec:diffmcq}).
    \item (\ref{sec:supp_ablations}) Additional experiments and ablations to supplement \refsec{sec:ablations}.
    \item (\ref{sec:supp_qual_results}) Qualitative results to add to those presented already in Figures~\ref{fig:extended_qa} and ~\ref{fig:diffcap_results}.
\end{itemize}

\section{Training data generation details} \label{sec:supp_data_gen}
As mentioned in \refsec{sec:dataset_gen}, we construct a paired QA dataset using pairs of video clips that share the same step label from HTStep~\cite{htstep}. In this section, we provide detailed descriptions of each phase in the data generation pipeline.

\paragraph{Action and object captioning} We use a VCLM model to describe actions and objects in the video clip~\cite{moon2023anymal} (see details in \refsec{sec:supp_implementation_details}). For actions, we sample 8 frames from the clip and use a HowTo100M~\cite{miech2019howto100m} trained captioning model. For object captions, we sample the center frame of the video clip and use an image captioning model~\cite{moon2023anymal}. The full prompt structure for each model is shown below

\begin{mdframed}[backgroundcolor=light-gray, roundcorner=10pt,leftmargin=0, rightmargin=0, innerleftmargin=4, innertopmargin=0, innerbottommargin=0, outerlinewidth=0, linecolor=light-gray]
\begin{lstlisting}[basicstyle=\LSTfont, breaklines=true, breakindent=0pt]
[SYSTEM PROMPT]
You are a multimodal assistant. Designed to provide direct answers to users' video related questions. Here is the video: {video}.

[ACTION PROMPT]
In one short sentence, describe what the person is doing?

[OBJECT PROMPT]
Give a very short list of all objects that are visible and their attributes, one per line. Only list objects being used, NOT in the background.
\end{lstlisting}
\end{mdframed} 

Despite the prompt asking to only list objects being used, the LLM-based captioning models tend to hallucinate object details that are not present in the scene. We therefore post-process the object captions using an off-the-shelf text grounding model~\cite{minderer2023scaling}. We retain only the object descriptions that have a grounding score greater than zero.

\paragraph{Consolidated step description} 

Next, we consolidate all the information above into a concise step description as shown in \reffig{fig:data_gen_pipeline} (left panel). For this, we use a text-only LLM model (Llama-2-70b-chat) with the following prompt.
\begin{mdframed}[backgroundcolor=light-gray, roundcorner=10pt,leftmargin=0, rightmargin=0, innerleftmargin=4, innertopmargin=0, innerbottommargin=0, outerlinewidth=0, linecolor=light-gray]
\begin{lstlisting}[basicstyle=\LSTfont, breaklines=true, breakindent=0pt]
[SYSTEM PROMPT]
You are an AI assistant that synthesizes the output of narration, action and object captioning models into a single description of the content.

[PROMPT]
Video narration: {narration}.
Possible activity: {action_caption}.
Possible objects: {object_caption}.
Summarize the captions into a single, descriptive sentence about what the person is doing, and using what objects.
\end{lstlisting}
\end{mdframed} 

\paragraph{Paired video QA generation} 

Finally, we select pairs of video clips, along with their generated step descriptions, and query the Llama-2 model to generate questions and answers. We generate questions of three types as shown below.
\begin{mdframed}[backgroundcolor=light-gray, roundcorner=10pt,leftmargin=0, rightmargin=0, innerleftmargin=4, innertopmargin=0, innerbottommargin=0, outerlinewidth=0, linecolor=light-gray]
\begin{lstlisting}[basicstyle=\LSTfont, breaklines=true, breakindent=0pt]
[SYSTEM PROMPT]
You are an AI assistant that asks questions comparing two videos based on their descriptions, and then answers them. Each question must be on a new line starting with "Q:" for question and "A:" for the answer. Use diverse language.
Video 1: {step_description_1}
Video 2: {step_description_2}

[PROMPT_TYPE1]
Summarize the differences and generate 3 question-answer pairs comparing the two videos. Answers should be short and concise.

[PROMPT_TYPE2]
Generate 3 question-answer pairs of the form "Which video ... ?". The answer must only refer to one of the two videos.

[PROMPT_TYPE3]
Do the two videos share a similar main action? Answer with a single word: YES or NO.
\end{lstlisting}
\end{mdframed}

\begin{figure*}
    \centering
    \includegraphics[width=\linewidth]{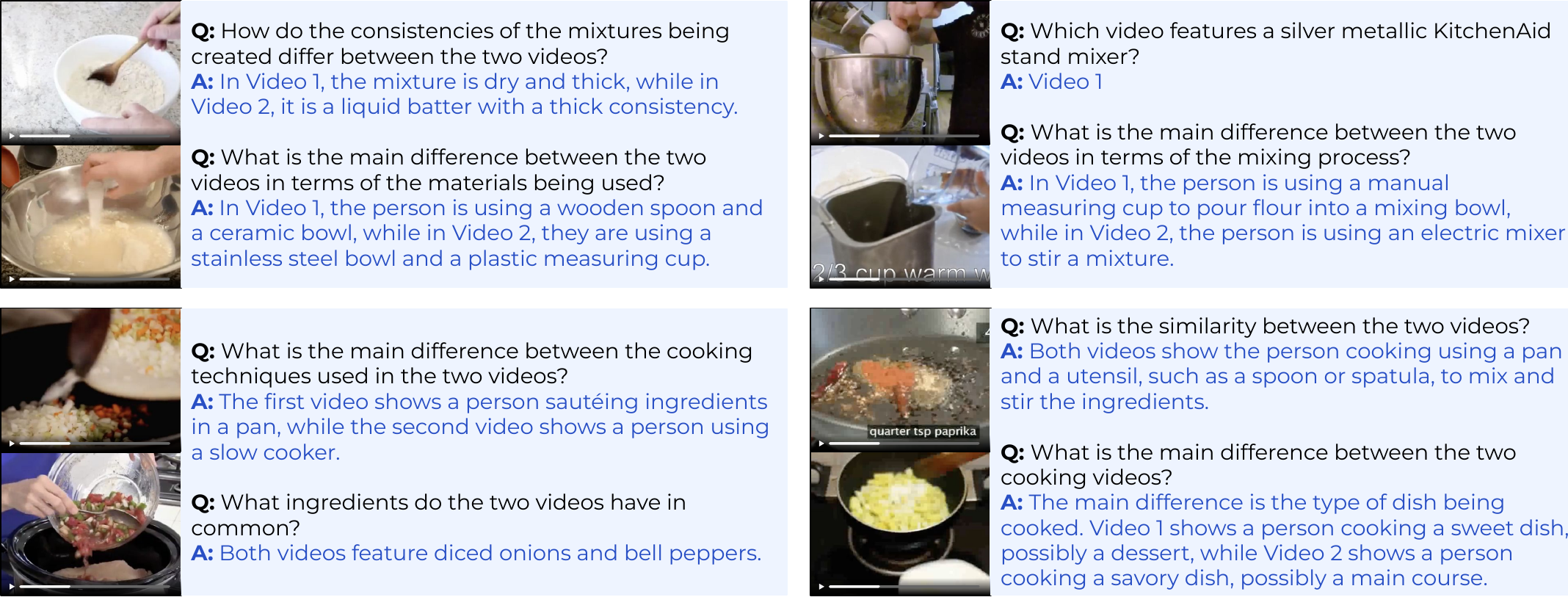}
    \caption{\textbf{Generated paired QA data.} Details in \refsec{sec:dataset_gen}. %
    }
    \label{fig:supp_pairqa_samples}
\end{figure*}

The final training dataset is the composition of question-answer pairs from all three sources. See \reffig{fig:supp_pairqa_samples} for examples of this data. Note that this data is used as weakly supervised training data only. For evaluation, a separate, disjoint set of video clips is manually annotated. See \refsec{sec:experiments} (dataset) and \refsec{sec:supp_annotation_details} for details.

\begin{figure*}[t]
\centering
\includegraphics[width=\linewidth]{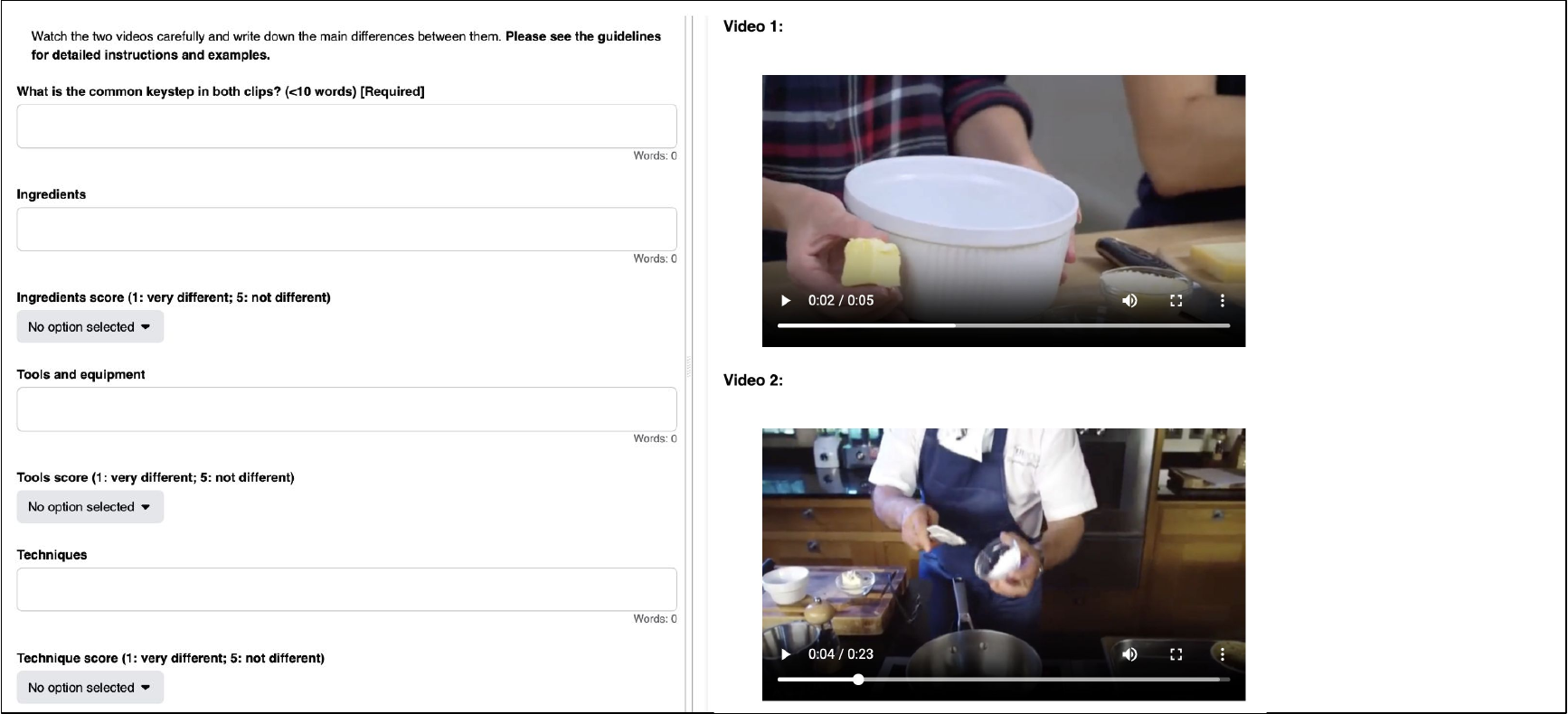}
\vspace{-0.2in}
\caption{\textbf{Data annotation interface} Annotators first watch two short video clips of a keystep performed by two different people (right panel). After that, they write out what they think the common keystep is between the two video clips, and then describe and score the differences between the clips them along various categories (left panel). Annotators can reject clips if they are not comparable (different keysteps, unclear or short videos).
}
\vspace{-0.1in}
\label{fig:supp_annotation_interface}
\end{figure*}

\begin{figure*}[t]
\centering
\includegraphics[width=\linewidth]{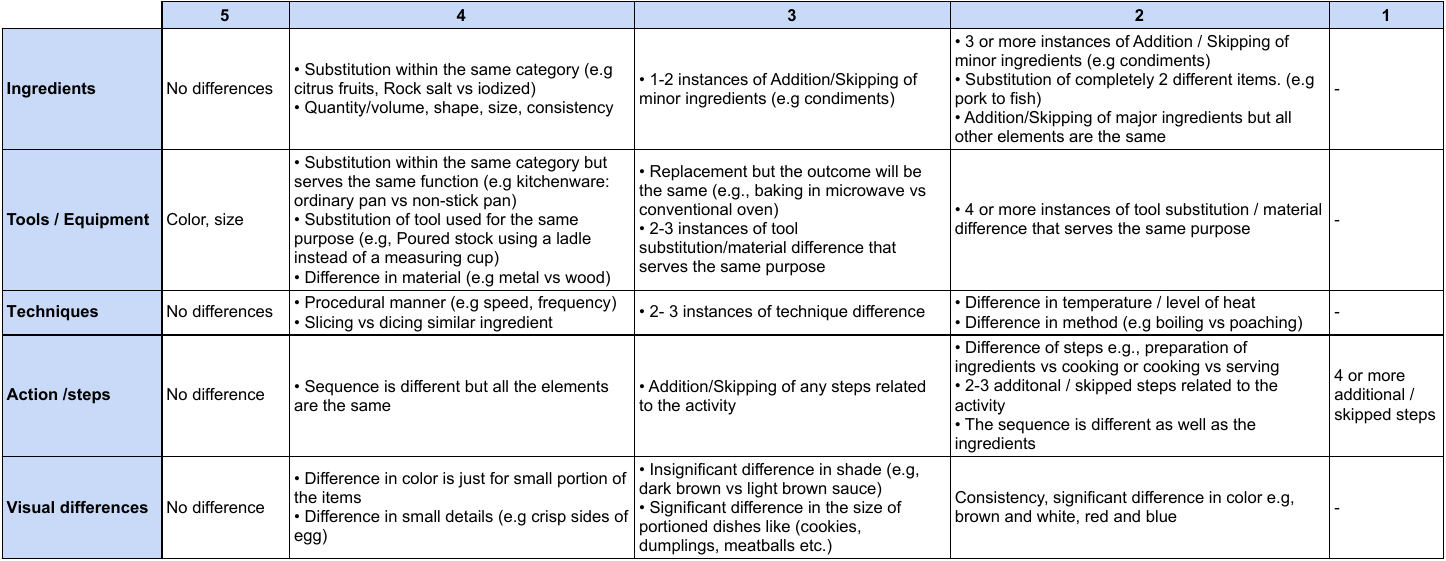}
\vspace{-0.2in}
\caption{\textbf{Difference scoring matrix} Annotators score how severe the differences are on a scale of 1-5 (1 = very different; 5 = nearly identical) using the scoring matrix as reference to avoid ambiguity across annotators.
}
\vspace{-0.1in}
\label{fig:supp_scoring_matrix}
\end{figure*}

\section{Annotation collection details} \label{sec:supp_annotation_details}
In this section, we provide details about the data annotation process outlined in \refsec{sec:experiments} (dataset). 

\paragraph{Annotation instructions and rubrics}
As mentioned in the main paper, annotators are presented with pairs of video clips from the same keystep category and asked to identify the main differences across 5 categories (ingredients, tools/equipment, techniques, visual differences) and then score how severe the differences per category are on a scale of 1-5. The annotation interface presented to the user is shown in \reffig{fig:supp_annotation_interface}. Scoring how severe the differences are is a fairly subjective task. To avoid ambiguity in this scoring, we present annotators with a scoring matrix (\reffig{fig:supp_scoring_matrix}) that provides a rubric for scoring differences in each category. We conducted pilot experiments to calculate inter-annotator agreement. We found that two out of three annotators agree 82\% of the time (Cohen's kappa = 0.64 on a [-1, 1] scale). Moreover, disagreements when present are small (on average within 1.2 points from each other).

\begin{figure*}[t]
\centering
\includegraphics[width=\linewidth]{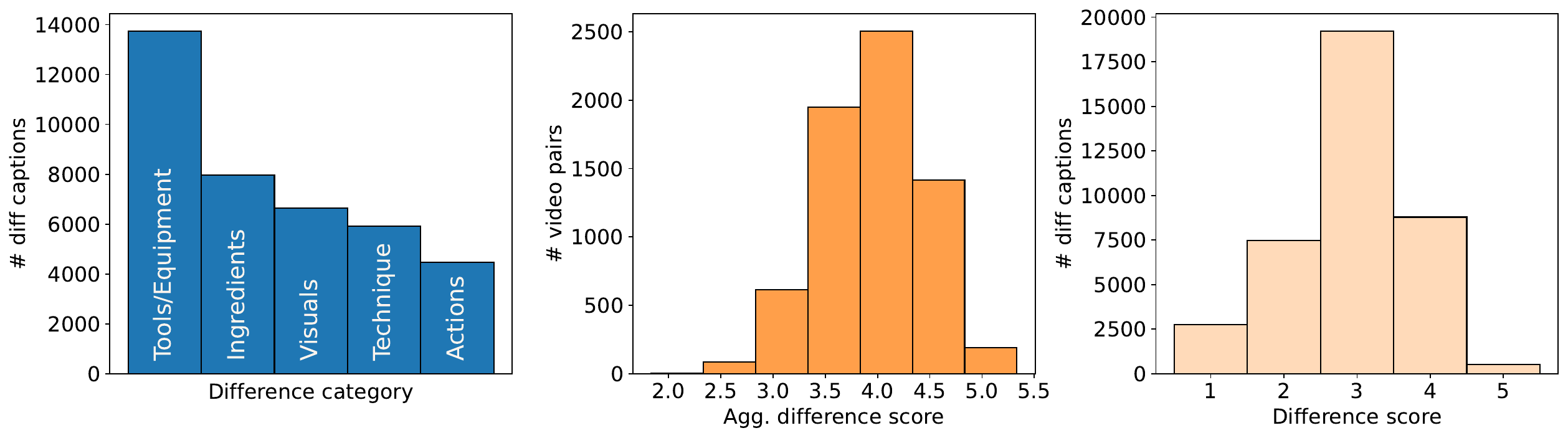}
\vspace{-0.2in}
\caption{\textbf{Annotated data statistics.} \textbf{Left:} Distribution of difference captions by category. \textbf{Middle:} Aggregate difference score distribution for video pairs (averaged over categories). \textbf{Right:} Distribution of difference scores for categories that have annotated differences (1 = very different; 5 = nearly identical).
}
\vspace{-0.1in}
\label{fig:supp_category_score_stats}
\end{figure*}

\paragraph{Dataset statistics and analysis}
Overall, we collect 35,988 difference captions across 6,292 video clip pairs involving 8,396 unique video clips. 
\reffig{fig:supp_category_score_stats} (left) shows the distribution of difference captions collected over the five categories, with \emph{Tools/Equipment} being the most popular category. There are fewer differences in \emph{Actions} which involves variations in step order, however they still account for a significant proportion of annotated differences (12\%). 
\reffig{fig:supp_category_score_stats} (middle) shows the aggregate difference score for video pairs in the dataset, computed by averaging the difference score across all categories. While all clip pairs are expected to be similar overall by design, since they are paired together if they share the same step label (on average, this aggregate score is 3.9), they often have significant differences in one or more individual category. \reffig{fig:supp_category_score_stats} (right) shows the distribution of difference scores only for categories where annotators label difference text, highlighting the spread in scores. 

\begin{figure*}[t]
\centering
\includegraphics[width=\linewidth]{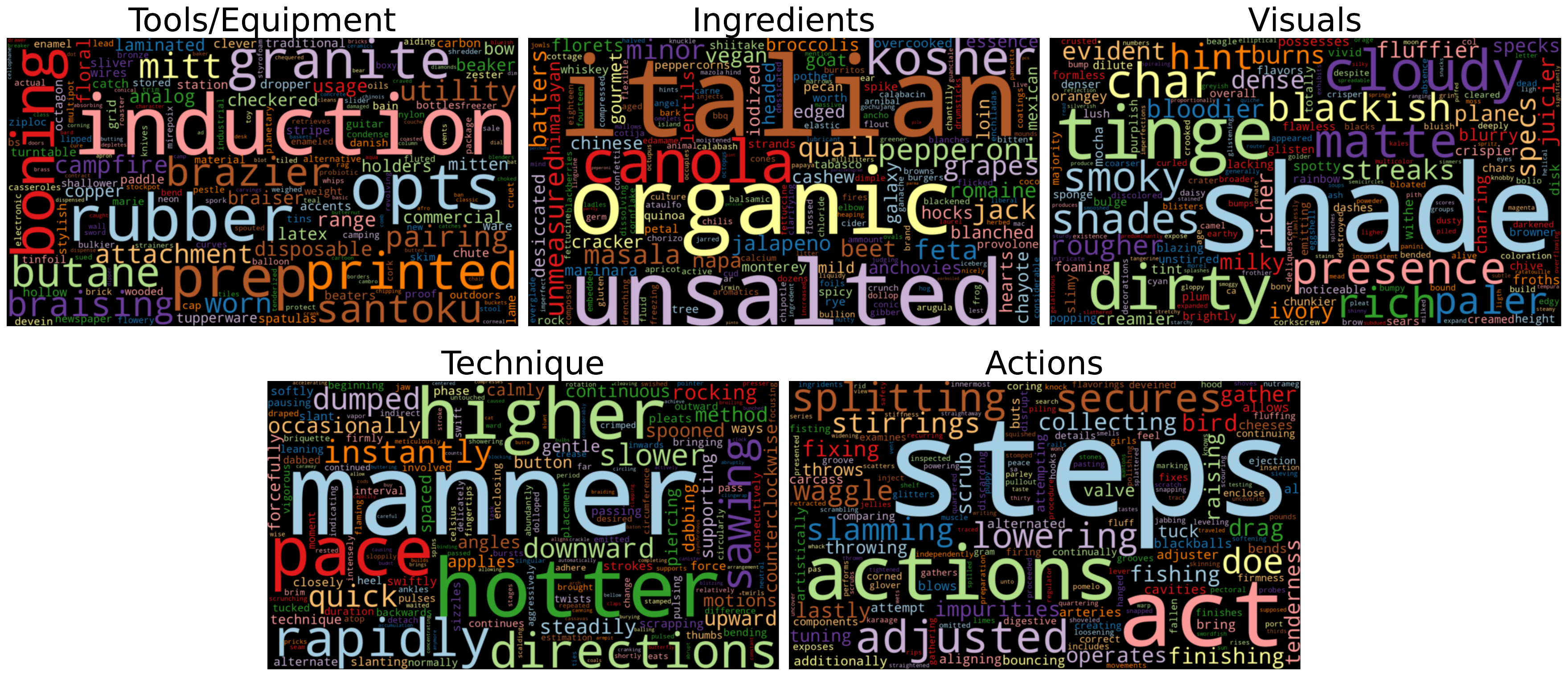}
\vspace{-0.2in}
\caption{\textbf{Prominent concepts captured in difference captions per category.} Tools/Equipment features tool materials and attributes (e.g., rubber, granite, butane), while techniques feature motion-related words (e.g., rapidly, quick, slowler).
}
\vspace{-0.1in}
\label{fig:supp_category_word_cloud}
\end{figure*}

In \reffig{fig:supp_category_word_cloud}, we show word clouds of prominent concepts captured in each difference category, sorted by their TF-IDF scores. We exclude words with a document frequency > 0.25 (e.g., person, instead, prefers etc.) to highlight category-specific concepts. We can see these concepts emerge for Tools/Equipment (e.g., materials, textures), Ingredients (e.g., ingredient names and properties), Visuals (e.g., visual attributes), Technique (e.g., motion-heavy words) and Actions (e.g., actions and verbs).

\begin{figure*}
    \centering
    \includegraphics[width=\linewidth]{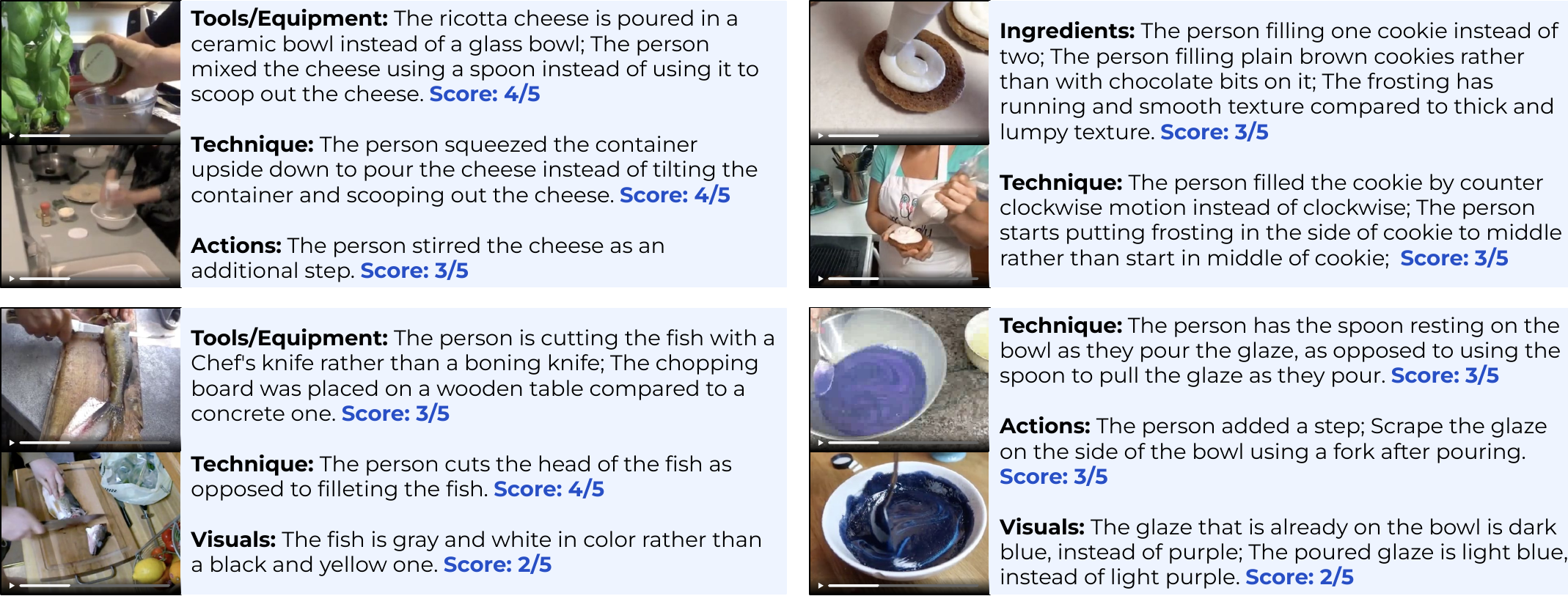}
    \caption{\textbf{Manually collected step differences.} Details in \refsec{sec:supp_annotation_details}. %
    }
    \label{fig:supp_stepdiff_samples}
\end{figure*}

Examples of these annotations can be seen in \reffig{fig:annot_examples} and \reffig{fig:supp_stepdiff_samples}. Note that none of these video clips are used in our automatic training data generation pipeline. These are a held-out subset of videos that are manually annotated for evaluation purposes only.

\section{Full implementation and training details} \label{sec:supp_implementation_details}
In this section, we present complete implementation details for our approach and all baselines listed in \refsec{sec:experiments}.

\paragraph{VCLM baselines} As mentioned in \refsec{sec:experiments} (baselines), we train our in-house VCLM and Interleaved baselines on clips from HowTo100M. To re-iterate, following prior work~\cite{moon2023anymal}, $M_V$ is an Internvideo~\cite{wang2022internvideo} video encoder that inputs 8 uniformly sampled frames from each video clip and generates 2056 spatio-temporal tokens. $M_{Proj}$ is a 2-layer Perceiver~\cite{jaegle2021perceiver} module followed by a linear layer head to output 32 tokens in the LLM's input dimension. During training, all parameters are frozen except for $M_{Proj}$.

For the VCLM models, we extract (video, ASR) pairs from automatically aligned ASR data from prior work~\cite{han2022temporal}. We use a batch size of 512 for 50k iterations. We use the AdamW optimizer, with a learning rate of 1e-4. For the Interleaved models, we sort (video, ASR) instances by their end timestamp and interleave sequences of 3 clips along with their ASR (clip1, ASR1, clip2, ASR2 ...). The Perceiver model converts each of the clips into 32 tokens. In addition to HowTo100M, we also train on single image captioning instances using filtered images from LAION2B~\cite{schuhmann2022laion} to improve the diversity of the training data beyond instructional video content. We duplicate the single image 8 times to feed to our video backbone. During training, we sample instances from each dataset in a round-robin manner. The batch size and number of iterations follow the VCLM models. 

\paragraph{StepDiff training details} As mentioned in \refsec{sec:experiments} (implementation details), we initialize our models from the Interleaved checkpoints above. In addition to LAION and HT100M data, we also train on our generated PairQA data from \refsec{sec:dataset_gen}. As before, we sample instances in a round-robin manner. We use a batch size of 256 for and train for 20k iterations based on validation data.

\section{Additional task formulation details} \label{sec:supp_tasks}

In \refsec{sec:downstream}, we described the prompts used for downstream tasks. To ensure that the outputs generated are in a consistent style with the collected annotations, we seed the generation step with partial text, and require the model to complete it. For DiffCap, we seed with ``The main difference in {category} is that in Video 2,'', and for DiffMCQ, we seed with ``In Video 2,'' followed by the difference caption text that is being evaluated. 

\begin{table}[t]
\centering
\small
\begin{tabular}{|l|ccc|}
\cline{2-4}
\multicolumn{1}{c|}{} & BLEU & CIDER & ROGUE-L  \\   \hline
Socratic (BLIP-2)~\cite{li2023blip}
& 0.122     & 0.016     & 0.139     \\
Socratic (LLaVA)~\cite{liu2023visual}
& 0.117     & 0.015     & 0.135     \\
Socratic (Step desc.)
& 0.113     & 0.009     & 0.139     \\ \cline{1-4}
VCLM (LLaVA)~\cite{liu2023visual}
& 0.143     & 0.037     & 0.144     \\
VCLM (AnyMAL)~\cite{moon2023anymal}
& 0.183     & \B{0.079} & 0.181     \\
Interleaved (IDEFICS)~\cite{laurenccon2023obelisc}
& 0.156	    & 0.041	    & 0.160     \\
Interleaved (AnyMAL)
& 0.184     & 0.068     & 0.185     \\ \cline{1-4}
StepDiff
& \B{0.193} & 0.061     & \B{0.191} \\ \hline
\end{tabular}
\vspace{-0.1in}
\caption{\textbf{DiffCap results without output parsing.} All methods perform worse on the generation metrics that are sensitive to sentence structure, though our method still has the best performance.}
\label{tbl:supp_diffcap_noparse}
\end{table}

Additionally, as mentioned in \refsec{sec:diffcap}, we post-process the outputs of each captioning baseline to match the annotated difference structure. This is important given the sensitivity of captioning metrics to even small structural changes. Even with careful prompting, the baselines tend to produce captions of the form ``In Video 1/2, the person ..., while in Video 2/1, ...'', while the annotations are collected in a specific format ``{action in candidate video} compared to {action in reference video}'' (see \reffig{fig:annot_examples}). The parsing involves simple text matching and replacing (e.g., replacing ``whereas in Video 1, the person'' with ``instead of''). Note that all models benefit from the same partial completion and output post-processing strategies listed above to ensure fair comparison.
In \reftbl{tbl:supp_diffcap_noparse} we show results without any additional parsing. All methods perform considerably worse compared to their counterparts with output parsing in \reftbl{tbl:results} (left), however our approach still achieves the highest performance among them. 

\section{Additional experiments} \label{sec:supp_ablations}
We present additional experiments to supplement the main paper results in \refsec{sec:experiments}. %

\begin{table}[t]
\centering
\small
\begin{tabular}{|l|cc|}
\cline{2-3}
\multicolumn{1}{c|}{} & CLIP~\cite{radford2021learning}  & InternVideo~\cite{wang2022internvideo}  \\   \hline
$V_r$ only      & 0.359     & 0.424     \\
$V_c$ only      & 0.353     & 0.413     \\  
$avg(V_r,V_c)$  & \B{0.396} & \B{0.451} \\   \hline
\end{tabular}
\vspace{-0.1in}
\caption{\textbf{VLEmbed variants.} Matching the difference caption to both the reference and the candidate video features results in the best performance.}
\label{tbl:supp_vlembed_alt}
\end{table}

\begin{table*}[]
\centering
\small
\begin{tabular}{|l|ccc|c|c|c|c|}
\multicolumn{1}{c}{} & \multicolumn{3}{c}{\SC{DiffCap}} & \multicolumn{1}{c}{} & \multicolumn{1}{c}{\SC{DiffMCQ}} & \multicolumn{1}{c}{} & \multicolumn{1}{c}{\SC{DiffRank}} \\
\cline{2-4}\cline{6-6}\cline{8-8}
\multicolumn{1}{c|}{}
& BLEU      & CIDER     & ROGUE-L   & & Acc \%      & & $\tau$      \\ \cline{1-4}\cline{6-6}\cline{8-8}
Socratic (BLIP-2)~\cite{li2023blip}
& 0.164     & 0.035     & 0.174     & & 0.341       & & 0.000       \\
Socratic (LLaVA)~\cite{liu2023visual}
& 0.155     & 0.027     & 0.169     & & 0.332       & & 0.000       \\
Socratic (Step desc.)
& 0.138     & 0.019     & 0.169     & & 0.400       & & 0.006       \\ \cline{1-4}\cline{6-6}\cline{8-8}
VCLM (LLaVA)~\cite{liu2023visual}
& \B{0.235} & 0.072     & 0.199     & & 0.385       & & 0.009       \\
VCLM (AnyMAL)~\cite{moon2023anymal}
& 0.193     & 0.106     & 0.196     & & 0.496       & & 0.041       \\
Interleaved (IDEFICS)~\cite{laurenccon2023obelisc}
& 0.187	    & 0.058     & 0.189     & & 0.340       & & 0.022       \\
Interleaved (AnyMAL)
& 0.221     & 0.105     & \B{0.216} & & 0.475       & & 0.048       \\ \cline{1-4}\cline{6-6}\cline{8-8}
StepDiff
& 0.216     & \B{0.124} & 0.205     & & \B{0.527}   & & \B{0.175}   \\ \cline{1-4}\cline{6-6}\cline{8-8}
\end{tabular}
\vspace{-0.1in}
\caption{\textbf{Results with lower capacity models.} Socratic (Llama 13B), AnyMAL (13B), LLaVA (7B) and IDEFICS (9B). Smaller models perform reasonably on the captioning task, but under-perform on the discriminative and ranking tasks.}
\label{tbl:supp_results_13B}
\end{table*}

\begin{table}[]
\centering
\small
\begin{tabular}{|l|ccc|}
\cline{2-4}
\multicolumn{1}{c|}{}
& V1      & V2     & V3 	\\ \cline{1-4}
VLEmbed (CLIP)~\cite{radford2021learning}
& 0.396       & 0.311       &  0.657         \\ 
VLEmbed (InternVideo)~\cite{wang2022internvideo}
& 0.451       & 0.336       &  \B{0.683}         \\ \cline{1-4}
Socratic (BLIP-2)~\cite{li2023blip}
& 0.335       & 0.219       &  0.644        \\
Socratic (LLaVA)~\cite{liu2023visual}
& 0.332       & 0.217       &  0.646        \\
Socratic (Step desc.)
& 0.392       & 0.258       &  0.648        \\ \cline{1-4}
VCLM (LLaVA)~\cite{liu2023visual}
& 0.381       & 0.319       &  0.561        \\
VCLM (AnyMAL)~\cite{moon2023anymal}
& 0.471       & 0.344       & 0.648         \\
Interleaved (IDEFICS)~\cite{laurenccon2023obelisc}
& 0.376       & 0.304       & 0.638         \\
Interleaved (AnyMAL)
& 0.497       & 0.351       & 0.653         \\ \cline{1-4}
StepDiff
& \B{0.541}   & \B{0.382}   & 0.654         \\ \cline{1-4}
\end{tabular}
\vspace{-0.1in}
\caption{\textbf{DiffMCQ variants for selecting negatives.} V1 excludes negatives that share the true reference or candidate video clip. This is the version reported in \reftbl{tbl:results}. V2 permits overlaps in reference / candidate clips as long as the pair is not identical. V3 fixes either the reference or candidate clip and randomly selects the other.}
\label{tbl:supp_diffmcq_alt}
\end{table}

\paragraph{Alternate variants of VLEmbed} In our experiments, we assumed that the embeddings of a \emph{pair of videos} can be represented as the average of their video embeddings. We evaluate other alternatives where a difference caption is matched to a single video (either the reference or the candidate) for DiffMCQ. Note that these variants are not applicable to DiffRank, where the difference caption is not an input. Our results in \reftbl{tbl:supp_vlembed_alt} show that including information from both video clips results in the best performance, though there is a small bias in the queries towards the reference video features.

\paragraph{Alternate variants of the DiffMCQ task} 
As mentioned in \refsec{sec:diffmcq}, we construct the task from the DiffCap annotations by sampling three \emph{negative} video pairs for every difference caption that are visually similar to the true video pair, but that do not exhibit the true difference. We identify the negatives as follows. First, we compute the average visual embedding (CLIP features) for each reference and candidate pair in the dataset, and sort the video pairs based on this distance to the positive pair embedding. Then, we go down this list and select pairs that obey two criteria: (1) they do not involve the true reference or candidate videos and (2) they do not share equivalent difference descriptions. For (2), we measure the sentence similarity between the ground truth difference and all of the differences for the selected pair in the category of interest, using MPNet~\cite{song2020mpnet} embeddings. If any difference text is too similar (above a threshold of $0.8$ cosine similarity), then we ignore the pair. We continue this process until we collect three negatives. 

Note that this is not the only method to construct the DiffMCQ task. For example, we can sample video pairs regardless of whether they share a reference or candidate video (as long as they are not the exact same pair). This results in a more difficult variant of DiffMCQ, but runs the risk of selecting negatives that may share differences. A third alternative is to fix either the reference or candidate clip and randomly sample the other, regardless of visual similarity or difference text similarity. We present all three alternatives in \reftbl{tbl:supp_diffmcq_alt}. Across the first two variants, our approach outperforms baselines. In the third alternative, the second clip is selected randomly, and so the VLEmbed baselines are sufficient for identifying outliers, and all baselines perform similarly. Moreover, the lack of constraints may permit negatives that still match the difference caption, making this version unsuitable for benchmarking our models.

\paragraph{Ablation experiments with lower capacity baselines} In \refsec{sec:ablations} of the main paper, we presented our method with a 13B parameter LLM backbone. In \reftbl{tbl:supp_results_13B}, we show results of all baseline models with smaller variants, including Socratic (LLama-13B), AnyMAL-13B, LLaVA-7B, and IDEFICS-9B. Our results show that while smaller capacity models perform reasonably well in the captioning task (even outperforming their 70B model alternatives on the BLEU metric), they perform worse overall on the discriminative and ranking tasks.

\section{Additional qualitative results} \label{sec:supp_qual_results}
\begin{figure*}
    \centering
    \includegraphics[width=0.9\linewidth]{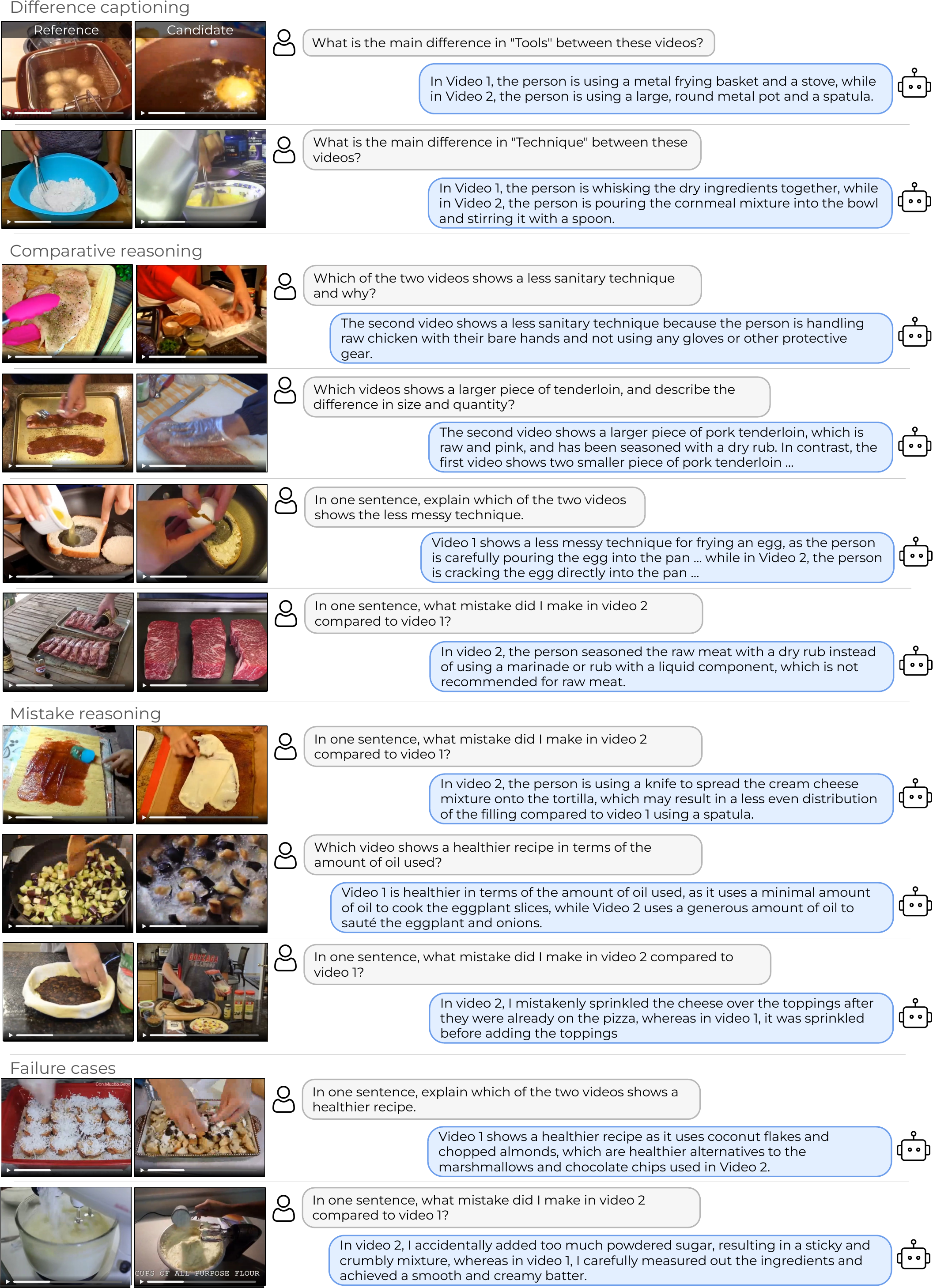}
    \caption{\textbf{Additional QA results on video pairs} See \refsec{sec:supp_qual_results} for discussion. Failure cases are shown in the last two rows. %
    }
    \label{fig:supp_extended_qa}
\end{figure*}

We show additional qualitative samples of our method's outputs in \reffig{fig:supp_extended_qa}. We show various kinds of supported prompts. These are standard difference captioning used to evaluate our models (panel 1), comparative reasoning (panel 2) and mistake reasoning (panel 3). Panel 4 highlights some failure cases. These typically arise due to two reasons. First, the underlying LLM naturally hallucinates details that are not present. This can happen due to inaccurate recognition (e.g., identifying a bell pepper as a jalapeno), or incomplete context information (e.g., without knowing the full recipe, the model assumes the dish is a dessert and the white powder is sugar). The second failure mode occurs when the model is forced to produce an output when differences in that category do not necessarily occur. This forces the model to hallucinate details as it is not trained to reject a query (e.g.,  asking ``what mistake did I make'' in the last row). More diverse automatically generated training data that explicitly handles these situations will likely address these failure modes. Despite these limitations, our approach can answer a wide variety of questions and requires reasoning over multiple videos, as shown in the figure.

\end{document}